# Robust tracking of respiratory rate in high-dynamic range scenes using mobile thermal imaging

YOUNGJUN CHO,[1,*] SIMON J. JULIER,[2] NICOLAI MARQUARDT,[1] AND NADIA BIANCHI-BERTHOUZE[1]

[1]*Interaction Centre, Faculty of Brain Sciences, University College London, London, WC1E 6BT, UK*
[2]*Department of Computer Science, University College London, London, WC1E 6BT, UK*
*\* youngjun.cho.15@ucl.ac.uk*

**Abstract:** The ability to monitor respiratory rate, one of the vital signs, is extremely important for medical treatment, healthcare and fitness sectors. In many situations, mobile methods, which allow users to undertake every day activities, are required. However, current monitoring systems can be obtrusive, requiring users to wear respiration belts or nasal probes. Alternatively, contactless digital image sensor based remote-photoplethysmography (PPG) can be used. However, remote PPG requires an ambient source of light, and does not work properly in dark places or under varying lighting conditions. Recent advances in thermographic systems have shrunk their size, weight and cost, to the point where it is possible to create smart-phone based respiration rate monitoring devices that are not affected by lighting conditions. However, mobile thermal imaging is challenged in scenes with high thermal dynamic ranges (e.g. due to the different environmental temperature distributions indoors and outdoors). This challenge is further amplified by general problems such as motion artifacts and low spatial resolution, leading to unreliable breathing signals. In this paper, we propose a novel and robust approach for respiration tracking which compensates for the negative effects of variations in the ambient temperature and motion artifacts and can accurately extract breathing rates in highly dynamic thermal scenes. The approach is based on tracking the nostril of the user and using local temperature variations to infer inhalation and exhalation cycles. It has three main contributions. The first is a novel *Optimal Quantization* technique which adaptively constructs a color mapping of absolute temperature to improve segmentation, classification and tracking. The second is the *Thermal Gradient Flow* method that computes thermal gradient magnitude maps to enhance accuracy of the nostril region tracking. Finally, we introduce the *Thermal Voxel* method to increase the reliability of the captured respiration signals compared to the traditional averaging method. We demonstrate the extreme robustness of our system to track the nostril-region and measure the respiratory rate by evaluating it during controlled respiration exercises in high thermal dynamic scenes (e.g. strong correlation (r=0.9987) with the ground truth from respiration-belt sensor). We also demonstrate how our algorithm outperformed standard algorithms in settings with different amount of environmental thermal changes and human motion. We open the tracked ROI sequences of the datasets collected for these studies (i.e. under both controlled and unconstrained real-world settings) to the community to foster work in this area.

## 1. Introduction

Monitoring respiratory rate plays a key role in a range of applications that span from direct diagnosis of, and treatment for, lung problems (e.g. hyperventilation, apnea and interstitial lung disease) and cardiovascular conditions to supporting a person's psychological needs (e.g. stress, anxiety regulation) [1,2]. Despite its importance, it has been largely disregarded in real world healthcare technology applications [3]. One possible reason is the inconvenience of conventional respiration measurement systems, such as chest-belts or oronasal probes, which demand direct physical contact [4–6]. These systems are often uncomfortable to wear and prone to motion artifacts which might cause incorrect sensor readings. In addition, in some medical and chronic conditions, where monitoring everyday physiological processes may be pivotal, direct contact with the skin may not be acceptable (e.g. Complex Regional Pain Syndrome [7]). Although contactless ways to measure respiration-related signatures (e.g. remote-photoplethysmography (PPG) [5,8], Doppler radar [9], thermal imaging [6,10–13] based measurements) can help address these limitations, they have not been investigated in the context of ubiquitous and mobile computing situations.

With the assistance of ambient lights (e.g. natural sunlight, lamp), a digital image sensor such as an RGB-based camera can be used as a remote PPG sensor [4,5,8,14,15] for monitoring blood volume pulse related parameters (e.g. heart rate). In [5,8], the authors found that PPG could also detect periodic respiratory periodic cycles in a known respiratory rate range (10-40 BPM in [5]), opening up the potential of real-time contactless breath tracking in stationary environments. [15] extended this work, showing that mobile sensing technology (a smartphone camera) could be employed as a remote PPG sensing channel. However, although the sensors for PPG are mobile, fundamental issues with PPG itself limit its flexibility and mobility. In particular, RGB-based cameras rely on moderate, stable ambient light levels. Therefore, they cannot easily support physiology monitoring in extreme lighting conditions which are either very dark or very bright. Furthermore, they struggle in situations where the light conditions continually change. As a result, most research with remote PPG has been applied in controlled conditions where the ambient lighting levels can be controlled (e.g. a constant 500 lux brightness in [4]). Other approaches can be used which are immune to ambient lighting conditions. Active sensors such as radar, for example, provide their own illumination. However, for respiratory rate tracking (such as [9]) they are restricted to monitor physiological parameters in stationary settings and indoors, where it is easier to ensure stillness of both the person and of the hardware installation, limiting their application in real-world deployments. On the other hand, thermal imaging does not have many of these constraints. Thermal imaging interprets the electromagnetic radiation which is naturally emitted from any object. Therefore, it can measure temperature in a passive way and does not require lighting sources. Furthermore, recent advances in commercial thermal imaging technologies have led to the development of new generations of thermal cameras (such as the FLIR One, Seek Thermal) which are relatively low cost (i.e. less than $250 in Jan 2017), lightweight, very small and can be connected to a smartphone. Therefore, it has been feasible to support real-time use in mobile situations. However, the use of mobile thermal imaging for physiology measurement has not been investigated, while large-sized, immobile thermal imaging systems have been employed demanding large physical spaces for the hardware installation similar to radar systems.

Work on thermography has shown that it is possible to track respiration in a contact-less manner by monitoring the temperature changes around the nostrils which are caused by inhalation and exhalation breathing cycles [6,10–13]. Most of these works define the nostril to be a key-region-of-interest (ROI) and have analyzed similarity between time-varying signals of the average temperature in the nostril-ROI to show inhalation and exhalation. This approach has been tested within several contexts in stationary and indoor settings where a thermal camera is installed, including neonatal care [13] and the study of breathing during sleep [12]. Despite its success, thermal technology is challenged by the difficulty of tracking ROIs in the presence of motion artifacts. Therefore, such issue must be addressed to improve accuracy of the breathing tracking. In very recent work on thermal imaging [6], the respiration tracking accuracy was greatly improved by using a latest automatic ROI-tracking method in computer vision (i.e. [16]) within indoor controlled experiments. Although state-of-the-art ROI tracking methods could handle motion artifacts, they do not provide robustness to environmental thermal dynamics. Even though our facial temperature distribution is internally controlled (e.g. blood vessel regulation), ambient temperature is an external factor that affects this thermal pattern. For this reason, the actual temperature within which signals can be found can vary. Such challenges are similar to the tone-mapping-related quantization issue found in converting real-world luminance to virtually expressed color [17]. Although lots of challenges have been identified in high dynamic range imaging, thermal imaging has entirely ignored the issue, adopting a fixed range of temperature (e.g. 28°C to 38°C). This has the potential to deteriorate the tracking performance under high thermal dynamic scenes. Other than the ROI tracking issue, the quality of respiratory feature is also influenced by motion artifacts and environmental thermal dynamics, which has not been broadly investigated by previous authors. Hence, a more robust respiration tracking approach is required to further improve the accuracy in mobile settings.

In this paper, we aim to address the three challenges (quantization issue in different ambient temperature, the ability to accurately track the nostril-ROI in more general conditions, enhancement of respiratory signal quality), through the development of a new approach for thermography-based breath tracking. In developing this approach, we make three main contributions: (1) a novel quantization technique which searches the optimal thermal range of interest on every single frame considering time-varying thermal dynamics, (2) an enhanced visual-tracking process of the nostril region containing in the presence of breathing dynamics, which is prone to shape changes amplified by motion artifacts, and (3) a novel signal representation based on a new concept of *thermal voxel* for extracting higher quality respiratory signals rather than just traditional ones (i.e. average temperatures on a ROI). The thermal voxel-based signatures are built on the fully tracked nostril-ROI sequences by means of the first two contributions (1,2).

Our proposed approach is thoroughly evaluated in both laboratory and real-world settings using low cost mobile thermography. Three different experiments were conducted: (i) controlled breathing patterns and motion artifact in four places with different ambient temperatures; (ii) unconstrained respiration with natural motion artifacts in indoor sedentary activity; (iii) unconstrained respiration in fully mobile contexts and with varying thermal dynamic range scenes (i.e. indoor and outdoor physical activity). In all cases, we show our algorithms were able to outperform existing approaches in the literature. Finally, we provide open access to the tracked ROI sequences of the datasets to foster future work in this area.

## 2. Background and key challenges

A key goal of this work is to formulate algorithmic methods which enable reliable monitoring of respiratory rate using mobile thermal imaging for a variety of health-related real-life applications. Until recently, thermographic systems have been regarded as infeasible and impractical solutions for physiological measurement because they were very costly and heavy [5]. The emergence of recently-launched mobile thermographic systems, which are low-cost and small-sized, opens the possibility of its use in this context. However, to have mobile, ubiquitous systems that operate in general environment and not merely controlled indoor-laboratory settings, we have to confront the following key challenges:

*Challenge 1: High thermal dynamic range scenes*

Despite our homoeothermic characteristics, temperature distribution on the cutaneous skin of the human face changes with the ambient temperature. Therefore, the changes in thermal patterns could lead to difficulties in identifying each facial area on thermal images. Figure 1(a), for instance, shows the effect of a person being exposed to ambient air while walking outside during the summer. As can be seen, as the person's face cools down, information about facial morphology becomes lost. However, this is a direct result of the way the image was created. The thermal camera measures temperature directly, and this must be quantized or converted into a digital signal with a fixed number of bits. The quantization process must cover where the bulk of the signal lies. The most common approach is to use linear quantization with a selected temperature range of interest, which is traditionally fixed from the first thermogram frame (e.g. 28°C to 38°C in [12]). However, this is unable to adapt to the dynamic situation here in which the temperature falls below 28°C in many parts of the image.

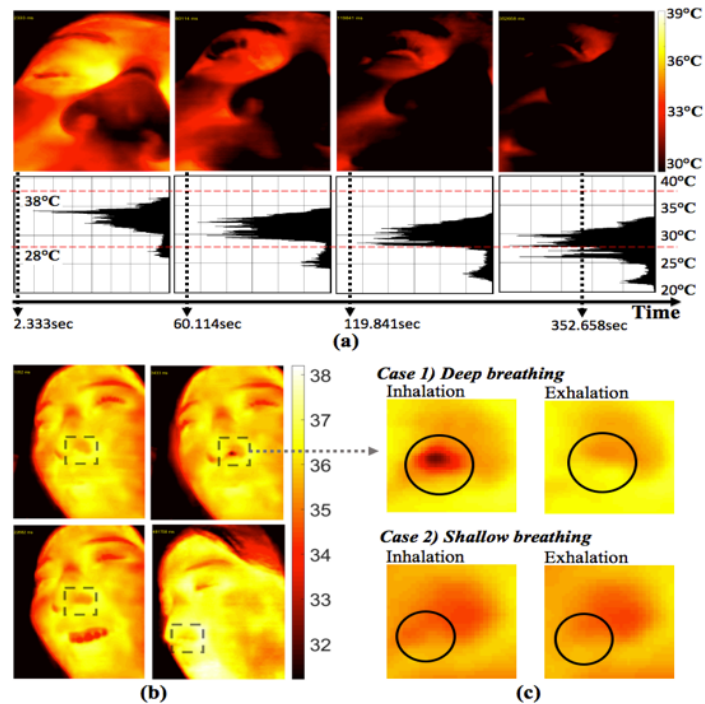

Fig. 1. Key challenges in thermal imaging-based respiration tracking. (a) high thermal dynamic range scenes: fixed thermal range of interest is not suitable in preserving the morphological facial shape within varying ambient temperature: [top] examples of thermogram shots collected from a person walking outdoor (for 6 minutes), [bottom] temperature histograms, (b) motion and breathing artifacts: the shape of the nostril is affected by mobility and respiration dynamics, (c) respiration signal quality: the four example shots show the tracked nostril region while breathing. The traditional average temperature is not ideal in extracting the respiratory feature when the respiration-induced thermal variance is weak, e.g. during shallow breathing in Case 2 compared with deep breathing in Case 1. The low spatial resolution of mobile thermal imaging also leads to the weak signal.

One way to address would be to increase the thermal range of values in color-mapping to ensure that all salient features are captured. However, because the resolution of the digital signal is fixed, this causes the resolution to go down. Because the range of temperature distribution over the facial area is relatively narrow, this would reduce contrast and the ability to detect subtle signals. Therefore, it is important to develop solutions to color mapping that can continuously adapt to take into account high dynamical thermal changes.

*Challenge 2: Combined motion artifact and air exchanges in the nostril*

A recent body of work on biomedical thermal imaging and remote-PPG has employed motion-tracking algorithms to extract physiological features from an ROI under motion. For example, [4,5] used the Lucas-Kanade (KLT) algorithm [18,19] to track facial areas where the PPG signal can be extracted. Pereira *et al.* [6] applied Mei *et al.* [16]'s Sparse Representation-based tracker. Although the body of work has achieved high performances of the tracking of ROIs in indoor constrained situations where just small amount of a person's head motion is allowed (e.g. [6]), unconstrained motion artifacts have not been generally tested. Furthermore, the thermally expressed nostril shape can undergo significant amount of deformations. These deformations result from motion and breathing dynamics. For example, the deformations shown in Figure 1(b) result from the participant turning their head and laughing. Therefore, although static contexts have been the main target for research, it is not possible to guarantee, in real life situations, that a person's nostril will be in a fixed, known shape. Therefore, new tracking algorithms must be developed which are robust to deformation of ROIs.

*Challenge 3: How to improve the quality of respiratory signals from a nostril ROI*
The most common way to track respiration rate is to analyze sequences of an average temperature in a tracked nostril region, which fluctuate caused by the expiration and inspiration cycles [6,10–13]. However, in many cases, the temperature change associated with breathing can be fairly small and affect only a small number of pixels. Figure 1(c), for example, illustrates the difference between shallow and deep breathing. Another difficulty is that the average temperature can be strongly affected by subtle location changes of the ROI bounding box and windy situations which can cause sudden global changes in the temperature distribution. A final difficulty is that when a person's viewing direction changes, it can also decrease the number of pixels that contain respiratory information. This is even more important for mobile thermal imaging systems, which often have low spatial resolution. Consequently, for the enhancement of the respiratory signal quality, there is a need to design a new feature which is robust to the various factors as discussed above.

## 3. Methods: robust respiration tracking algorithm

To overcome the challenges mentioned above, we propose the approach illustrated in Figure 2. The approach has three main components: (i) *Optimal Quantization* of the mapping of temperature elements to digitalized color-mapped pixels that can adapt to different thermal dynamic range scenes, (ii) *Thermal Gradient Flow* – a nostril tracking algorithm to reduce effects of motion and breathing dynamics, and (iii) *Thermal Voxel-based Respiratory Rate Estimation* to enhance the respiratory signal quality.

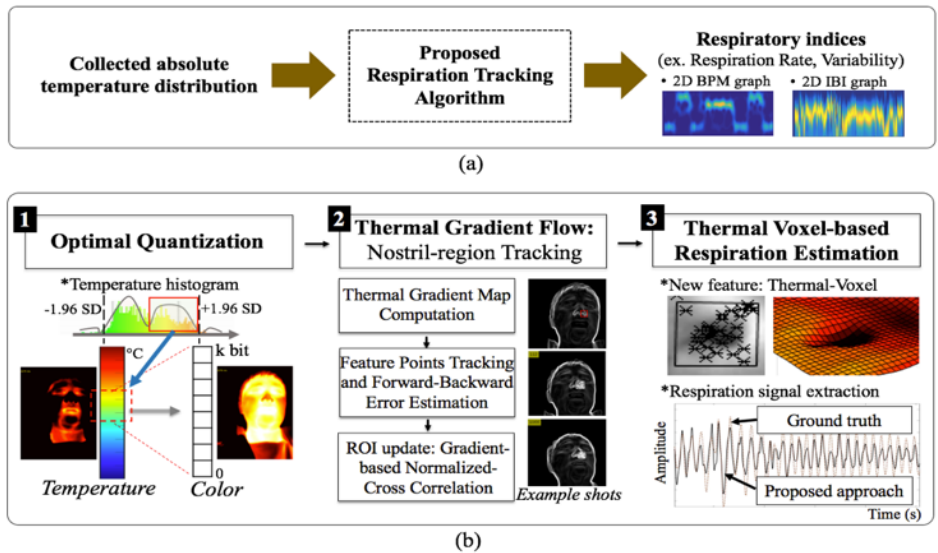

Fig. 2. (a) Overall procedure of the imaging-based respiration tracking system; (b) key components: 1) Optimal Quantization – convert from the absolute temperature distributions to the color-mapped images by analyzing the temperature histogram of every frame, 2) Thermal Gradient Flow – nostril-region tracking method using the thermal gradient magnitude and points tracking methods, and 3) Thermal Voxel-based Respiratory Rate Estimation – extracting the respiratory signals by integrating the unit thermal voxels inside the nostril.

*3.1 Optimal Quantization for high thermal dynamic range scenes*

In thermal image processing, quantization is the process of translating from a continuous temperature value to its digital color-mapped equivalent. Suppose the temperature lies in the range $[T_0, T_{k-1}]$ and the color pixel which represents it lies in the range $[u_0, u_{k-1}]$. We define $u = \Gamma(T)$ as the mapping between the two. For the quantization in time-varying thermal dynamic range scenes, our idea is to adaptively quantize the thermal distribution sequences by finding a thermal range of interest that contains the whole facial temperature distributions for every single frame (see Figure 2(b)-1). As a first step, we use a statistical extreme value removal process to reduce unexpected noises (e.g. sunlight projected on a person's glasses) and points of extreme temperature, which may be produced by errors of mobile thermal imaging in calculating temperature due to fixed emissivity or lens conditions (e.g. misted lens). By removing thermal signals beyond 1.96 standard deviations, we can set an initial candidate for the thermal range $[T'_{min}, T'_{max}]$:

$$T'_{min} = \bar{c} - 1.96 \frac{\sigma_c}{\sqrt{n \cdot m}}, \quad T'_{max} = \bar{c} + 1.96 \frac{\sigma_c}{\sqrt{n \cdot m}} \quad (1)$$

where $\bar{c}$ is the sample mean of $c(x)$ which is the one-dimensional temperature distribution ($\forall x \in 1, n \cdot m$), $n \cdot m$ is the spatial resolution of collected thermal distribution matrixes, and $\sigma_c$ is the standard deviation of $c(x)$.

Now, for the final selection of the range of interest, we assume there are two qualitatively different elements in each frame: the person's face, and the background. Therefore, we adopt Ridler and Calvard's concept of *optimal threshold selection* [20]. This method finds the threshold value that best separates objects from background by iteratively analyzing the color histogram. This can help to search thermal values that distinguish the human skin from non-skin areas from the time-varying temperature histograms exhibiting various dynamic ranges. Furthermore, it is known that this method is robust even in the presence of non-bimodal histograms which make it harder to find an optimal boundary of temperature ranges of interest. Our Optimal Quantization technique, which is named after the concept, is finalized with the iterative computation of an optimal threshold value $T_{opt}$ below:

$$T_{opt}(0) = T'_{min} \quad (2)$$

$$T_{opt}(t+1) = \frac{\mu_1(t) + \mu_2(t)}{2} \quad (3)$$

where $\mu_1(t)$ and $\mu_2(t)$ are the mean values when $c(x) \leq T_{opt}(t)$, $c(x) > T_{opt}(t)$, respectively. While the method presented in [20] requires the four corners of an image to contain background pixels, our method does not require this condition because Eq. (2) is initialized with a value from Eq. (1). The process is iterated until $T_{opt}(p) - T_{opt}(p-1) \approx 0$ is satisfied and the temperature range of interest is then chosen to be

$$T_0 = T_{opt}(p), \quad T_{k-1} = T'_{max}. \quad (4)$$

When the average temperature over the background, which includes hair and air, is lower than that of human cutaneous skin, only the lower bound is of interest in determining the optimal range. When the average temperature of the background is above that of human cutaneous skin, the upper bound can be used instead.

*3.2 Thermal Gradient Flow: Nostril-region tracking*

To achieve robust performance in the tracking of the nostril, we propose a new algorithm called *Thermal Gradient Flow*. Given the color-mapped images produced through Optimal

Quantization, this algorithm computes the thermal gradient magnitude matrix for each frame and employs Kalal *et al.*'s Median Flow algorithm [21] which uses forward-backward error estimation on points tracked by Lucas-Kanade's disparity-based tracker [19]. To enhance robustness, we compensate for the loss of feature points by resetting a ROI based on the gradient-based two-dimensional normalized cross correlation, as illustrated in Figure 2(b)-2.

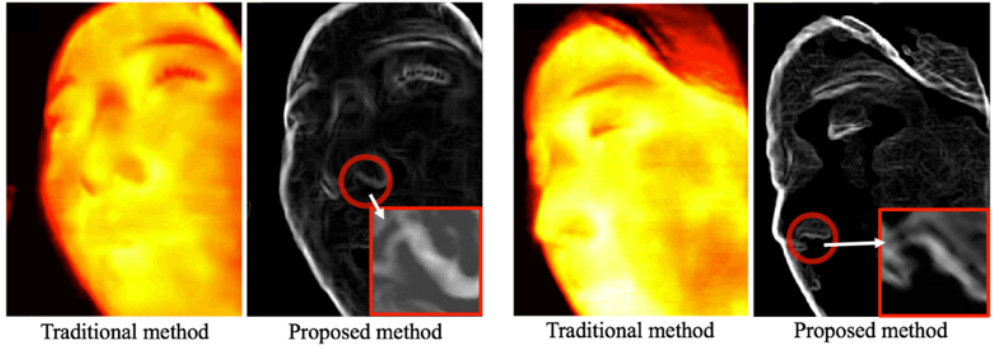

Fig. 3. Example shots of conversion from thermal images to thermal-gradient magnitude maps: the proposed method can help to preserve the morphology of the nostril region during motion (Zoomed-in-areas are manually rotated for the visual representation).

The human homeothermic metabolism and the relatively low thermal conductivity of the human skin act as low pass filters. As a result, the shape of nasal and nostril areas is often blurred, leading to a weak differentiation between key facial point features. To obtain clearer features, we enhance the boundary between the nostril and the ala of the nose by converting the quantized thermal image $u$ into the two-dimensional thermal-gradient magnitude map $\Phi$ by

$$\Phi(x,y) = \sqrt{\left(\frac{\partial u(x,y)}{\partial x}\right)^2 + \left(\frac{\partial u(x,y)}{\partial y}\right)^2} \qquad (5)$$

where $x$ and $y$ are the coordinates in the x-y plane of the image space. As illustrated in Figure 3, the more distinct morphological shape of the nostril can be obtained from the thermal-gradient based image than the normal thermal image under combined artifacts (i.e. motion and respiration dynamics). The converted image can then be used to collect the feature points representing the nostril for the motion-tracking.

The thermal-gradient map of the nostril-ROI, $\Phi_{ROI}$, is chosen by selecting the nostril as an ROI of size of $N \times N$ pixels in the first frame. This can be done manually by a human or automatically (e.g. as in [6]). As our point tracker, we use the *Median Flow* algorithm [21] because it has proven its tracking performance on thermal imaging in a non-biomedical sector [22]. This algorithm calculates the forward-backward error defined as

$$e(T_f^k | S) = \|x_t - \hat{x}_t\| \qquad (6)$$

where the thermal-gradient image sequence is $S = (\Phi_t, \Phi_{t+1}, ..., \Phi_{t+k})$, the forward trajectory is $T_f^k = (x_t, x_{t+1}, ..., x_{t+k})$ and the backward trajectory $T_b^k = (\hat{x}_{t+k}, \hat{x}_{t+k-1}, ..., \hat{x}_t)$ produced by backward tracking up to the first frame. Here, $\hat{x}_{t+k} = x_{t+k}$ and $\|x_t - \hat{x}_t\|$ is the Euclidean distance between the two points. In this algorithm, the points are tracked using the KLT [19].

In our case, the points are selected from the nostril ROI on the thermal-gradient magnitude map. For more details on the implementation, we refer to [21].

As a final strategy to handle the case when the point features are completely lost, we can use two-dimensional normalized-cross correlation [23]. In particular, the combination of correlation with gradient has been shown to give high performance in the registration of deformable components in neuroimaging [24,25]. Similarly, we enhance the tracking performance of the nostril by searching for a new position of the ROI which maximizes the *gradient-based normalized-cross correlation coefficient*

$$\gamma(\mathbf{x}) = \frac{\sum_i (\Phi_{ROI}(\mathbf{x}_i) - \mu_{\Phi_{ROI}(\mathbf{x}_i)})(\Phi(\mathbf{x}_i - \mathbf{x}) - \mu_{\Phi(\mathbf{x}_i - \mathbf{x})})}{\sqrt{\sum_i (\Phi_{ROI}(\mathbf{x}_i) - \mu_{\Phi_{ROI}(\mathbf{x}_i)})^2 \sum_i (\Phi(\mathbf{x}_i - \mathbf{x}) - \mu_{\Phi(\mathbf{x}_i - \mathbf{x})})^2}} \qquad (7)$$

where $\mathbf{x}$ is a set of $(x,y)$ in (5) at the center of $N \times N$ square (i.e. same with the size of a ROI). When the number of tracked points falls below a certain threshold, this method resets the ROI and finds new gradient-based point features. This method can also be applied to the automated ROI selection at the first frame in case we have a plenty of nostril-image sets.

*3.3 Respiratory rate estimation through Thermal Voxel integration*

Heat exchange in nostrils during inhalation and exhalation is determined by a person's breathing pattern. To monitor these patterns, existing methods most frequently compute the average value of spatial thermal distribution inside the nostril ROI on every single frame [6,10–13] and seek for fluctuating patterns over time. However, as explained earlier, there are many issues with using the average of the distribution. Therefore, in this section, we present a new three-dimensional *Thermal Voxel*-based feature to further enhance the quality of the breathing signals. Inspired by the use of voxels in neuroimaging [26], the approach maps each two-dimensional unit thermal into a voxel three dimensional space (see Figure 4(a,b)). This feature can be immune to factors inducing low quality of breathing information and global changes of spatial thermal distribution since it focuses on extracting breathing-induced thermal volume changes inside the nostril by computing the quantity of inside thermal voxels (see Figure 4(b)). It is constructed from

$$\hat{v}(t) = \int_{T_{\min}(t)}^{T_\delta(t)} \Lambda_t(T) dT \approx \sum_i \sum_j T_\delta(t) - \hat{u}_{ij}(t), \quad T_\delta(t) - \hat{u}_{ij}(t) > 0 \qquad (8)$$

where $\Lambda_t(T)$ is the integral of the thermal voxels in the nostril in a cross section with the temperature $T$, $\hat{u}_{ij}$ is the absolute temperature on the tracked region and $T_\delta$ is the upper boundary to integrate the concave volume which is set to a temporal moving average (here, n=2) of the spatial mean temperature values to not only have a stable boundary but to consider global thermal changes. If the ROI tracker loses track of the nostril region, $T_\delta$ would need to be reset from the next frame to reject its value from the misplaced bounding box. The misplacement could be detected by inspecting a sudden change (i.e. differential) of statistical skewness on the thermal distribution:

$$\Delta \gamma_1^t = E\left[\frac{\hat{u}_{ij}(t) - \mu_t}{\sigma_t}\right]^3 - E\left[\frac{\hat{u}_{ij}(t-1) - \mu_{t-1}}{\sigma_{t-1}}\right]^3. \qquad (9)$$

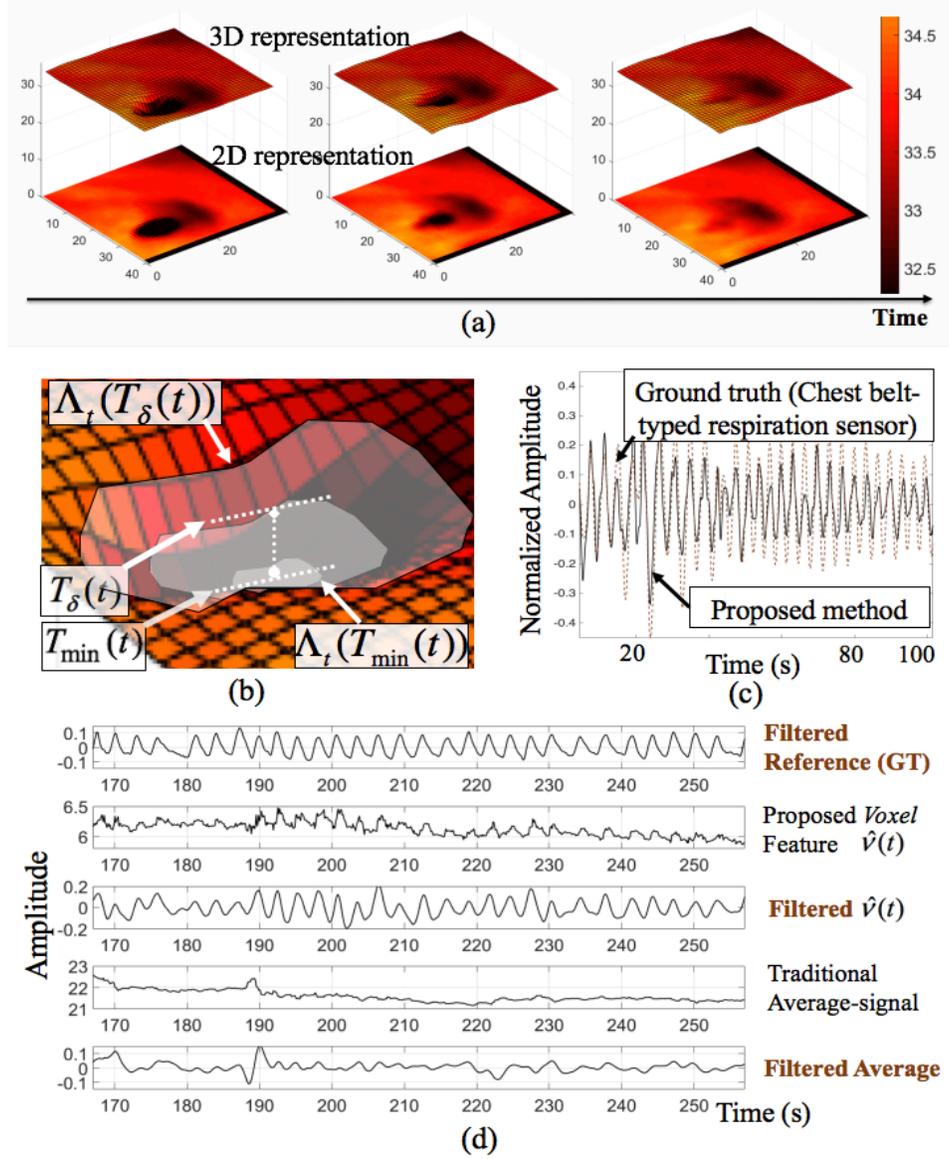

Fig. 4. Extraction of respiratory patterns through Thermal Voxel integration: (a) a person's nostril and its thermogram sequences along the time in 3D (top) and 2D (bottom), (b) the concave volume corresponding to heat variances in the nostril, (c) the extracted respiratory signals compared with ground truth signals, and (d) a comparison of the filtered *voxel*-based signals with the traditional method as the participant changes their head. The voxel method closely tracks ground truth, but the traditional method fails.

To estimate respiratory rate, both frequency domain (such as peak frequency detection [5]) and time-domain (such as the Bayesian approach based on short-time estimators [27] employed in [6]) approaches could be used. In this paper, we used the short-time power spectral density, which analyzes the self-similarity of the thermal voxel feature $\hat{v}(t)$ to determine the rate. This is achieved by computing the Fourier transform $F_f$ of the short-time autocorrelation function [28]. To decrease ripples in the frequency domain due to the truncated short-time window, we use the Gaussian window

$$w_i(k) = \hat{v}(n_i + k)e^{-\frac{1}{2}\left(\frac{k}{\sigma}\right)^2}, \quad k \in \{-\hat{t}_{\max}f_s, \ldots, \hat{t}_{\max}f_s\}. \tag{10}$$

The window length $w_i(k)$ is $2\hat{t}_{\max}f_s + 1$, where $\hat{t}_{\max}$ is the upper time limit of respiration and $f_s$ is the sampling frequency. The value of $\hat{t}_{\max}$ is determined by the expected minimum respiratory rate of interest. For our experiments, we use the same range of expected breathing rate as in [6] of 0.1Hz to 0.85Hz. Therefore, $\hat{t}_{\max}$ = 10s. Once $w_i(k)$ has been computed, its value is normalized by feature scaling. The output is filtered through a third order elliptic filter (a passband ripple of 3 dB and a stopband attenuation of 6 dB) with passband cutoff frequency of 0.1Hz and 0.85Hz. Given this, we estimate the respiratory rate by searching for frequency $f$ which maximizes the power spectral density

$$S_V(f) = F_f(R_{ww}) = \sum_k R_{ww}(k)e^{-j2\pi fk} \tag{11}$$

where $R_{ww}$ is the short-time autocorrelation of the filtered $w_i(k)$.

## 4. Datasets and experimental protocols

In this section, we describe the experimental protocols which were designed to gather thermal imaging of respiration data of different levels of complexity in terms of thermal dynamics and motion artifacts for an in depth evaluation of our proposed methods. The fully tracked nostril ROI sequences from the collected datasets are opened to the research community.

### 4.1 Experimental setup

For both the laboratory and the real world settings, we used a mobile thermal imaging camera (FLIR One for Android - dimensions: 72mm x 26mm x 18mm, FLIR Systems Inc., Santa Barbara, CA, www.flir.com). This device detects electromagnetic waves in the spectral range of 8 to 14μm with a spatial resolution of 160x120 and a temporal resolution of less than 9 frames per second (fps). The thermal images were recorded by using recording software (Thermal Camera RRT, Android Play Store). To improve the accuracy of the temperature measurement, it is necessary to consider the emissivity coefficient of the surface targeted. In our study, the emissivity is set to 0.98 that is the emissivity value of the human skin [29].

To evaluate the performance of our approach, a commercial respiration sensor was used. Participants were required to wear a chest belt-based respiration sensor (ProComp Infiniti Resp/SA9311M, Thought Technology). This reference respiration sensor produces the respiration waveform by monitoring expansion and contraction of the chest or the abdomen. The reference sensor collects the data at 256Hz. Therefore, to allow a direct comparison, we up-sampled the data sequences from the Thermal Voxel-based method with spline interpolation to 256Hz.

The proposed algorithmic methods were implemented in MATLAB (2015b, The MathWorks). The validation process was carried out on a 64-bit Windows 7 desktop (Core i3-4160T 3.10GHz process, Intel) with 8GB RAM.

### 4.2 Procedure and dataset

The experimental protocols were designed with different scenarios along with higher complexity reflecting the thermal dynamic ranges and motion artifacts. Each experiment was independently designed for different purposes and each dataset was collected from a new group of participants. The aim of the first experiment protocol is to generate a dataset for a systematic evaluation of the method in different thermal dynamic range scenes by using guided breathing patterns and by constraining the person's movement as in Gastel *et al.*'s [5]. Both the second and third experiment protocols are designed to evaluate the proposed methods under natural breathing conditions: the second one targets sedentary activities without constraining movement but with controlled ambient temperature; the last experiment targets physical activity in fully mobile contexts with varying ambient temperature. Following Gastel *et al.*'s [5], only 5 participants were invited for the first experiment protocol. Given the very controlled nature of this experiment, this number was considered sufficient to ensure the robustness of the test. For the other two experiments, a slightly higher number was used since higher variability between participants breathing patterns was expected. Since, due to ethics restrictions, only mild physical activity was used at this stage and only healthy participants were invited such slightly higher number was considered sufficient to cover the expected amount of variability.

The researcher was in charge of running the experiments. Prior to each experiment, participants were asked to read the information sheet and sign the informed consent form. The experimental protocols were approved by the Ethics Committee of University College London Interaction Centre. The different experiments (as described below) led to the creation of three datasets. The ROI sequences, which were completely tracked by Optimal Quantization and Thermal Gradient Flow, from all the datasets are publicly available at *http://youngjuncho.com/datasets/*.

*4.2.1. Dataset 1: controlled respiration in environments with non-constant temperature*

The aim of the first experiment was to carry out a systematic evaluation of our approach in environments with different temperature values and dynamics. 5 healthy adults (2 female) (aged 29-38 years, M=31.4, SD=3.78) were recruited from the university subject pool. Following the protocol used in [5], participants were asked to maintain a stable posture and breath according to a set of breathing patterns presented to them on a screen. Figure 5 shows the design for this experiment. As shown in Figure 5(a), all the participants were given a thermal camera attached to an Android smartphone to record their face and an additional smartphone that provided the breathing patterns. Figure 5(b) shows the three guiding breathing patterns composed of slow (10 breaths/min), normal (15 bpm) and fast speed (30 bpm). Each breathing pattern lasted 30 seconds. The guiding breathing patterns were displayed dynamically on the screen. Participants were given a 60 seconds-training period. Taking advantage of mobile thermal imaging, participants were able to monitor themselves by aiming the camera at their face. The distance between the face and device ranged from 35cm to 55cm. The recordings were repeated in four different places: a controlled room ("*Place A*"), entrance of the building (wind from outside and heat from inside) ("*Place B*"), a street corner (windy) ("*Place C*") and park ("*Place D*") in winter. The collected dataset consists of approximately 80 minutes - recordings (5 participants x 4 places x 4 minutes). The person was asked to remain as still as possible.

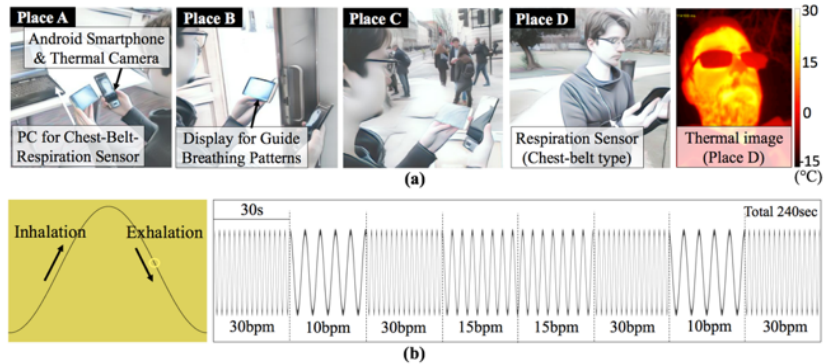

Fig. 5. Experiment 1: (a) to obtain different thermal dynamic range scenes (i.e. environments with non-constant dynamic temperature), four different places were chosen (Place A: room, B: entrance of the building, C: corner on the street, D: park), the last image-shot is a thermal image collected in Place D, the experiment was run in winter, (b) the guiding breathing patterns are composed of three different rates (10(slow), 15(normal), 30(fast) breaths/min).

*4.2.2. Dataset 2: unconstrained respiration during desk activity with natural motion artifacts*

The aim of the second experiment was to test our approach in more realistic unconstrained sedentary desk activities. 10 healthy adults (6 female) (aged 24-31 years, M=28.4, SD=2.17) from a variety of ethnical backgrounds (skin colour: from pale white to black) were recruited from the university subject pool. The pool includes people from outside the university. The experiment was conducted in a quiet laboratory room in summer and simulated desk activity behaviour, consisting of three phases lasting 2 minutes each: i) *sitting and conversation,* ii) *reading a news article on the screen* and iii) *surfing the internet with the keyboard and mouse*. As described in Figure 6(a), the mobile thermal camera was installed near each participant's face using a shoulder rig and the distance from the face ranged from 35cm to 50cm to account for the spatial resolution (160x120) of the camera. To ensure natural motion artefacts, people were told to act naturally, i.e. no movement constraints were imposed. The collected dataset includes a variety of movement situations, such as head rotations due to people walking behind them with temporary disappearance of the nostril from the thermal camera view. The change in a participant's position from phase i) to ii) ensured changes in global temperature variance around the nostril-ROI. The experiment resulted in 60 minutes (10 participants x 6 minutes) of thermal video recording of spontaneous breathing patterns, natural movements in sedentary contexts and changes in ambient temperature.

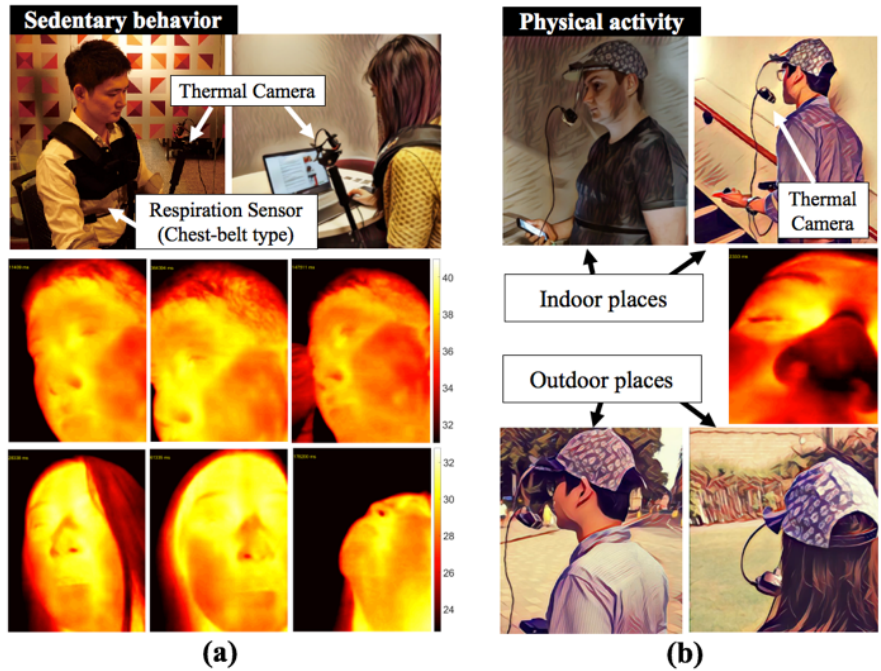

Fig. 6. (a) experiment 2: unconstrained respiration in desk activities, (b) experiment 3: unconstrained respiration in both indoor and outdoor light physical activities.

*4.2.3. Dataset 3: unconstrained respiration in fully mobile context and varying thermal dynamic range scenes*

The last experiment aimed to measure the respiration patterns for people undertaking natural, unrestricted actions. In order to enable mobility, the thermal camera was attached to a headset-microphone-shaped rig whose distance from the face ranged between 20cm and 30cm as shown in Figure 6(b). We recruited 8 healthy adults (5 female), aged 23-31 years (M=27.0, SD=2.93) from various ethnical backgrounds. To simulate a variety of fully unconstrained situations, the experiment had two main sessions: i) indoor physical activity and ii) outdoor physical activity. The first session consisted of three tasks of 2 minutes each: walking through a corridor, standing in a dark room while doing small movement and climbing and descending stairs. The second session was carried out outdoor on a street pavement and in a windy park to involve varying thermal dynamic range scenes. During the session, subjects were guided to walk slow, walk fast, and stroll in natural paces. Each walking pattern lasted 2 minutes. Example shots are shown in Figure 6(b). All sessions were run in the summer. The final dataset includes thermal imaging sequences of approximately 96 minutes (8 participants x 2 sessions x 3 activities x 2 minutes).

*4.3 Evaluation process*

*4.3.1. Evaluating ROI-tracking performances*

Three state of the art visual tracking algorithms proposed between in 1990s and 2010s were implemented to evaluate the ROI-tracking performance of the Optimal Quantization based Thermal Gradient Flow method. First, Mode-Seeking (also called mean-shift) algorithm [30] was selected since it is known as a simple and effective traditional method. For a comparison with one of the latest work in thermal imaging-based tracking of respiratory rate [6], Mei *et al.*'s Sparse Representation [16] was chosen as one of the most advanced methods in the area. Lastly, Kalal *et al.*'s Median Flow [21] was implemented as it has been recently used in thermal imaging [22] and also forms the backbone of our proposed Thermal Gradient Flow method. Since the employed methods do not handle the dynamic quantization issue, the thermal range of interest to ±5°C from the average temperature over a person's whole face in the first thermogram frame. This fixed range was used for static quantization as in [12] (i.e. the range is [28°C, 38°C]). In most cases, the algorithms stopped working when a tracking fault occurred, i.e. tracker going off beyond 50% of the nostril region. We also manually checked the tracking faults to confirm the number of fully tracked frames. This number was used as tracking performance. For the parameter settings for both the Median Flow and the Thermal Gradient Flow, the maximum backward-forward error allowed (i.e. Eq. (6)) was set to 5 for Datasets 1 and 2 (i.e. 'small' ROI), and to 25 for Dataset 3 (i.e. 'big' ROI). The setting of the values was also based on the physical distance between the camera and the nostril. A repeated measures analysis of variance (ANOVA) test was used to investigate the effects of the ROI-tracking algorithms on tracking performances. In addition, in the case of Dataset 1, the effect of environment (i.e. indoor vs. outdoor places) was also considered. Hence, we first conducted a two-ways repeated measures ANOVA to investigate the effect of each variable on the tracking performances. Then, to better compare the performances of the four algorithms, two separate one-way repeated ANOVA tests were carried out, one for the indoor (i.e. indoor room Place A) and one for outdoor data (i.e. outdoor places: Place B-D). Given that the environments did not vary substantially in the case of Datasets 2 and 3, only the effect of type of algorithm was investigated using one-way repeated measures ANOVA tests. Finally, for each datasets, the ANOVA tests were followed by pairwise post-hoc Bonferroni tests with adjustment [31] to compare the performances of the different algorithms.

*4.3.2. Evaluating respiration rate tracking performances*

To evaluate the tracking of the respiratory rate, the Thermal Voxel based respiration estimation method was compared with the traditional and main approach of using the average temperature over the nostril ROI [6,10–13]. In this evaluation, we used the ROI-sequences that were automatically tracked by Optimal Quantization-based Thermal Gradient Flow tracking method. As reference signals, the waveforms collected from the chest-belt-respiration sensor were used as described in Section 4.1. To enable the comparison of the two sensed signals, these were automatically synchronized using the *Maximum-Amplitude of Cross-Correlation (MACC) alignment* expressed as $\max(R_{fg}(\eta))$ where $R_{fg}$ is the cross-correlation between the reference signals and the estimated signals, and $\eta$ is the discrete lag. As shown in Figure 7(a), we synchronized the two signals by analyzing periodic similarity.

Even though periodicity changes throughout the experiment, correct alignment could be achieved when the cross correlation is applied across all the data for a single trial. To examine statistical agreement levels between two different respiratory rate measurements, we used the Bland-Altman plot and the root mean square error (RMSE) as in [4]. Here, the length of the time-window (i.e. Eq. (10)) was set to 20 seconds with 15 seconds overlap (i.e. 75%).

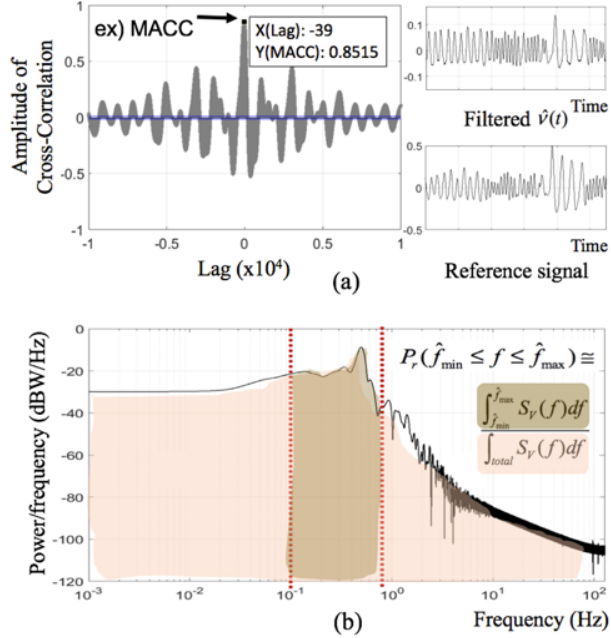

Fig. 7. Statistical methods for evaluations: (a) automated-synchronization between estimated signals and reference signals using the *maximum-amplitude of cross-correlation (MACC)*, (b) respiration-related goodness probability as a respiratory signal quality index (rSQI).

*4.4 Respiratory signal quality index (rSQI)*

To examine the respiration signal quality, we use a respiratory Signal Quality Index (rSQI) which helps to identify moments when the extracted signals are of poor quality due to malfunctioning of the sensor. We adapted the goodness metric concept introduced by Kumar *et al.* [4], which analyzed the power level of physiological signs in frequency ranges of interest (e.g. breathing rate: 0.1Hz to 0.85Hz used in [6]) to assess the quality of extracted signals. Both contact and non-contact respiration sensors are known to be influenced by motion artefacts. For example, the accuracy in measuring the respiration pattern is affected by physical activity. In the case of the chest-belt sensor, the movement of a person (e.g. while walking) may lead to changes in the belt position and hence to its tension. We extend the goodness metric proposed in [4] by dividing the band of interest by the total energy within a half of the sampling frequency (e.g. here 128Hz from 256Hz) to satisfy a statistical probability condition. As a rSQI, we define the respiration-related goodness probability $P_r$ as

$$P_r(\hat{f}_{\min} \leq f \leq \hat{f}_{\max}) \cong \frac{\int_{\hat{f}_{\min}}^{\hat{f}_{\max}} S_V(f)df}{\int_{total} S_V(f)df} \quad (12)$$

where $0 \leq P_r \leq 1$, $S_V$ is the power spectral density in Eq. (11), and $\hat{f}_{\min}, \hat{f}_{\max}$ are the lower (e.g. 0.1Hz) and upper (e.g. 0.85Hz) boundaries of the expected breathing rate, respectively.

Figure 7(b) shows the concept and the area between the dotted lines is the numerator of Eq. (12).

## 5. Results

### 5.1 Nostril-region tracking performance

We assessed the performance of each tracking method (Mode Seeking [30], Sparse Representation [16], Median Flow [21] and Thermal Gradient Flow which is based on Optimal Quantization) by computing the percentage of frames which were successfully tracked over all the trials. Figure 8 summarizes the results for the nostril-region tracking in *Dataset 1* (controlled respiration in environments with thermal dynamic changes), *Dataset 2* (unconstrained respiration during sedentary activity) and *Dataset 3* (unconstrained respiration during physical activity). Figure 9 compares the level of motion artifacts by using a Euclidian distance (unit: pixel) between the origin of a ROI in the first frame and that in the current frame. Compared with those from the controlled conditions (Figure 9(a)), the sedentary activity (Dataset 2) and physical activity (Dataset 3) produced the higher levels of motion artifacts. Furthermore, different types of motion artifacts were found according to the type of activity: high peaks of the tracker's movement due to change of one's head direction during the sedentary activity and high fluctuations of the movement due to the oscillation of the rig during the physical activity.

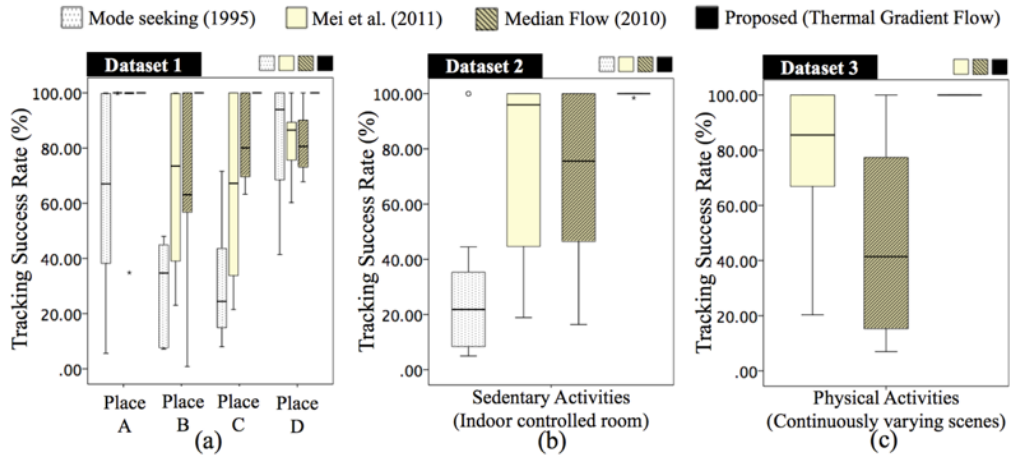

Fig. 8. Overall results of the nostril-tracking performance of Thermal Gradient Flow compared with existing methods: (a) Dataset 1 (controlled but in non-constant temperature scenes), (b) Dataset 2 (unconstrained respiration during sedentary activity), (c) Dataset 3 (unconstrained respiration during physical activity).

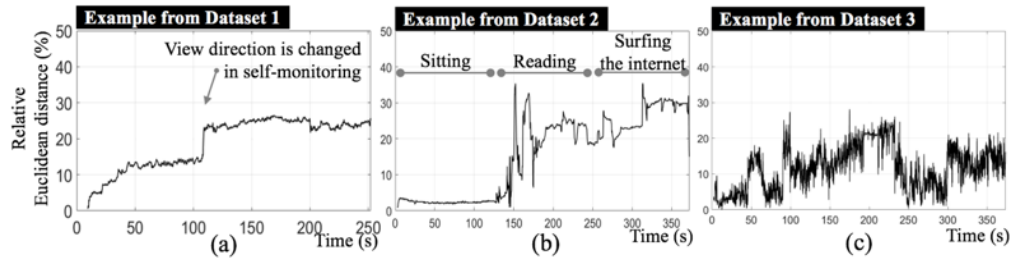

Fig. 9. Quantified motion artifacts using the relative Euclidean distance from the origin of the nostril-ROI at the first frame: (a) example from Dataset 1 (fully controlled), (b) from Dataset 2 (sedentary behavior), and (c) from Dataset 3 (physical activity).

*Dataset 1: Overall results*

For the data collected at *Place A* (a room, which has a *low thermal dynamic* range), while Mode Seeking performed the worst, the other three produced highly reliable results similarly (see Figure 8(a) - Place A). On the other hand, for the data collected in *Places B-D* (i.e. outside) characterized by *high thermal dynamic* ranges, Thermal Gradient Flow produced significantly better results and was able to track all frames in all situations (Mode-seeking: M=47.25%, SD=33.04, Sparse Representation: M=71.31%, SD=29.37, Median Flow: M=76.35%, SD=26.16, and Thermal Gradient Flow: M=100.00%, SD=0.0) as summarized in Figure 8(a).

As described in Section 4.3.1, we tested the overall success rates using two-way repeated measures ANOVA. Although we found that the type of environment ($F(3,12)=6.881$, $p<0.01$), the algorithm ($F(3,12)=7.602$, $p<0.005$) and the interaction between the two independent variables ($F(9,36)=2.24$, $p<0.05$) had significant effects on the correct success rate as summarized in Table 1, no significant differences within each variable were found from the pairwise post-hoc analysis.

Table 1. Effects of different environments (i.e. different thermal dynamic range scenes) and algorithms on the success rate of the nostril-region tracking from Dataset 1 using two-way repeated measures ANOVA

| Source | Sum of squares | df | Mean square | F | p-value |
|---|---|---|---|---|---|
| Environment | 7778.596 | 3 | 2592.865 | 6.881 | 0.006 |
| Algorithm | 24255.468 | 3 | 8085.156 | 7.602 | 0.004 |
| Environment*Algorithm | 7061.431 | 9 | 784.603 | 2.24 | 0.042 |

Then, we grouped Place B-D as outdoor condition, and separately tested significance of each ROI tracking methods for outdoor (i.e. high thermal dynamics) and for indoor (i.e. low thermal dynamics) conditions, using one-way repeated measures ANOVA. Interestingly, there was a stronger significant effect of the type of algorithm on the success rate in the outdoor (i.e. Place B-D) condition ($F(3,42)=12.237$, $p<0.001$) as described in Table 2. The pairwise comparisons for this outdoor condition showed significant difference between the performance of Thermal Gradient Flow and those of Mode-Seeking ($p=0.000$) and Sparse Representation ($p=0.012$). Although there was no significant difference between Thermal Gradient Flow and Median Flow ($p=0.021$), Thermal Gradient Flow provided higher results (see Figure 8(a) - Place B-D). On the other hand, there was no significant differences between the performances of the tracking algorithms in the indoor condition ($F(3,12)=2.828$, $p>0.05$) as summarized in Table 3. This result is consistent with the findings from the literature where high performances of the ROI tracking in indoor environments are reported (such as [6]), indicating that state-of-the-art methods were reliably implemented for our experimental evaluation.

Table 2. Significance test to assess effects of different algorithms on the success rate in the outdoor condition (i.e. Place B-D in Dataset 1, *high thermal dynamics*) using one-way repeated measures ANOVA

| Source | Sum of squares | df | Mean square | F | p-value |
|---|---|---|---|---|---|
| Algorithm | 21060.252 | 3 | 7020.084 | 12.237 | 0.000 |

Table 3. Significance test to assess effects of different algorithms on the success rate in the indoor condition (i.e. Place A in Dataset 1, *low thermal dynamics*) using one-way repeated measures ANOVA

| Source | Sum of squares | df | Mean square | F | p-value |
|---|---|---|---|---|---|
| Algorithm | 4777.893 | 3 | 1592.631 | 2.828 | 0.083 |

*Datasets 2 and 3: Overall results*

For data collected while the persons were on the move (i.e. Datasets 2 and 3), our approach outperformed all the other methods (Dataset 2 – Figure 8 (b): Mode-seeking: M=28.47%, SD=28.37, Sparse Representation: M=76.70%, SD=31.58, Median Flow: M=67.95%, SD=34.65, and Thermal Gradient Flow: M=99.84%, SD=0.49; Dataset 3 – Figure 8 (c): Sparse Representation: M=78.21%, SD=26.86, Median Flow: M=48.19%, SD=35.02, and Thermal Gradient Flow: M=100.0%, SD=0.0). Considering the low performance of Mode-Seeking found in Datasets 1 and 2, we excluded this method in the comparisons for Dataset 3. Given the independent variable in Datasets 2 and 3 is the type of algorithm, a one-way repeated measures ANOVA test was used. The results are summarized in Table 4 and 5. Here, strong significant effects of algorithm on success rate results were found with $F(3,27)=17.822$ ($p<0.001$) for Dataset 2 and $F(2,30)=17.029$ ($p<0.001$) for Dataset 3. The pairwise post-hoc analysis confirmed that Thermal Gradient Flow produced significantly better results than Median Flow ($p=0.000$) and Sparse Representation ($p=0.016$) in the contexts of both high thermal dynamic ranges and motion artifacts. In the case of motion artifacts only (Dataset 2), no significant differences were found between Thermal Gradient Flow and the advanced tracking methods. Once again, we found that Mode-Seeking performed worst and Thermal Gradient Flow performed best (Sparse Representation: $p=0.016$, Median Flow: $p=0.031$, Thermal Gradient Flow: $p=0.000$). By all accounts, our method appears to be the most robust in high dynamic temperature range scenes.

Table 4. Effect of different algorithms on ROI tracking performance in sedentary activities (i.e. Dataset 2) using one-way repeated measures ANOVA

| Source | Sum of squares | df | Mean square | F | p-value |
|---|---|---|---|---|---|
| Algorithm | 26522.909 | 3 | 8840.970 | 17.822 | 0.000 |

Table 5. Effect of different algorithms on ROI tracking during physical activities (i.e. Dataset 3) using one-way repeated measures ANOVA

| Source | Sum of squares | df | Mean square | F | p-value |
|---|---|---|---|---|---|
| Algorithm | 21654.360 | 2 | 10827.180 | 17.029 | 0.000 |

*5.2 Respiratory rate estimation performance*

Figure 10 compares Thermal Voxel-based respiration estimation method with the temperature averaging approach to extracting respiratory signals. Figure 10(a, d, g) shows, in the time domain, the respiratory signals extracted from both methods and the ground truth (belt-sensor) for a subject randomly chosen as an example: (a) for subject S5 from Dataset 1, (d) for subject S10 from Dataset 2, and (g) for subject S7 from Dataset 3. Figure 10(b, e, h) illustrate the spectrograms (produced using RVS in [34]) showing the calculated respiration rates corresponding to the same subjects. The respiration-related goodness probability $P_r$ (see Eq. (12)) from each time-window was computed and the overall data from every subject is described in the histogram chart (see Figure 10(c, f, i)). From the sequences for Dataset 1, the $P_r$ of each measurement shows similar distributions for the three methods (Thermal Voxel-based: M=0.9895, SD=0.0103, average-based: M=0.9815, SD=0.0181, ground-truth: M=0.9825, SD=0.0174), indicating that the respiration rates calculated by each measurement were generally reliable. For the sequences for Datasets 2 and 3, on the other hand, different $P_r$ distributions were found. Overall, lower signal quality levels were observed in comparison with the controlled situations (i.e. Dataset 1). More specifically, for Dataset 2 (i.e. sedentary activity), while Thermal Voxel-based method and the ground truth had a similar distribution (Thermal Voxel-based: M=0.9639, SD=0.0468, ground-truth: M=0.9694, SD=0.0437), the quality of signals extracted by the traditional average-based method appeared to be more deteriorated (M=0.9287, SD=0.0771). From the spectrogram, for example, patterns resulted from Thermal Voxel integration method are more similar to those from the ground truth in comparison with the traditional method (e.g. see the label * in Figure 10(e)). In the case of Dataset 3 (physical activity), our method produced the highest signal quality while the ground truth from the belt-sensor suffered more from the physical movement (Thermal Voxel-based: M=0.9640, SD=0.0434, average-based: M=0.9555, SD=0.0490, ground-truth: M=0.9170, SD=0.0964). In particular, except for the segment involving a stationary task (i.e. labeled as *A* in Figure 10(g,h): standing with a small movement in a dark room where there were less motion artifacts), the ground truth method produced relatively unclear patterns in the spectrogram.

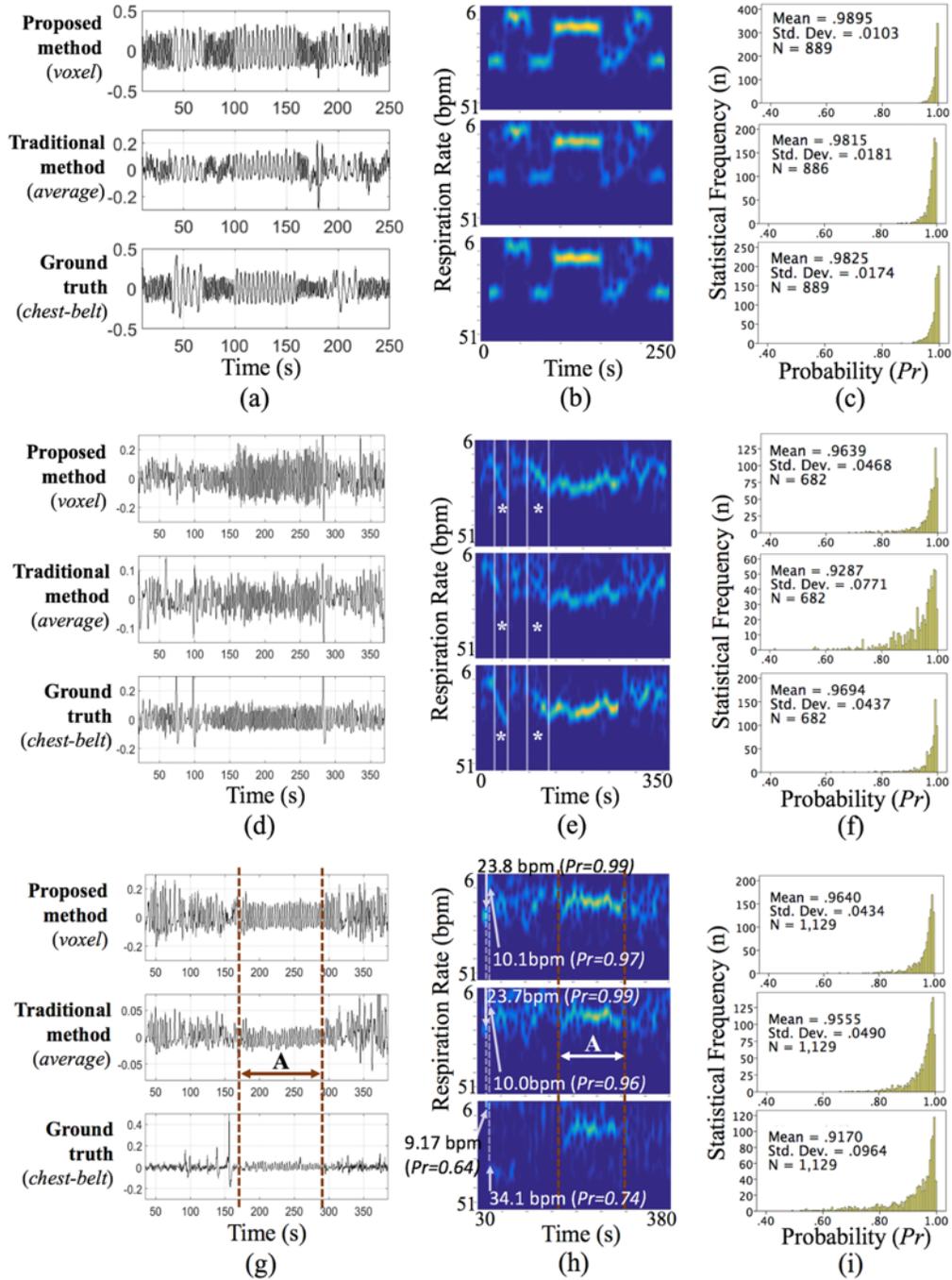

Fig. 10. Results of respiratory signature extraction: (a,d,g) time-domain signal, (b,e,h) frequency-domain signal (respiration rate), and (c,f,i) the respiration-related goodness metric. The Thermal Voxel-based method is more robust than the traditional averaging-based method for Dataset 1 (a-c) and Dataset 2 (d-f). For Dataset 3 (g-i) (i.e. fully mobile contexts), the ground truth shows less reliability in the respiration tracking, except for the segment A in (g) (i.e. standing with small movement).

*Dataset 1: Bland-Altman and RMSE analysis*

Figure 11 summarizes the overall accuracy results of the respiratory rate estimation for Dataset 1. The Thermal Voxel integration method produced highly robust performances: mean bias of 0.0882 bpm with the 95% limits of agreement being -0.7956 to 0.9721 bpm (Figure 11(a)). By contrast, the traditional averaging method produced the mean bias of 0.0755 bpm with the 95% limits of agreement being -1.8670 to 2.0179 bpm (Figure 11(b)). In addition, the Thermal Voxel-based method shows stronger correlations with the ground truth ($r=0.9987$, $p<0.001$) from the belt sensor than the traditional averaging method ($r=0.9936$, $p<0.001$) does. As summarized in Figure 11(c), the RMSE of Thermal Voxel integration method (0.459bpm) was lower than the error of the traditional averaging method (0.993bpm). Finally, we compare the results of the methods separately over each type of environment provided in Dataset 1. Figure 12 summarizes the results of the analysis for different environment in Dataset 1. In particular, the performance of the traditional method tends to be more affected by the different range of thermal dynamics in comparison with our method (RMSEs of Thermal Voxel integration method: 0.452bpm for Place A, 0.525bpm for Place B, 0.420bpm for Place C, 0.429bpm for Place D, RMSEs of temperature averaging method: 0.651bpm for Place A, 1.286bpm for Place B, 1.181bpm for Place C, 0.686bpm for Place D).

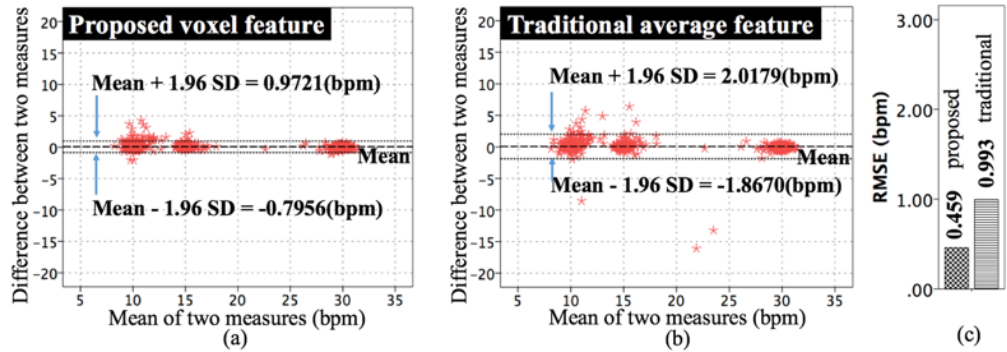

Fig. 11. Dataset 1 (overall): Bland-Altman plots of (a) Thermal Voxel integration method, (b) the traditional averaging method, and (c) overall RMSE comparisons.

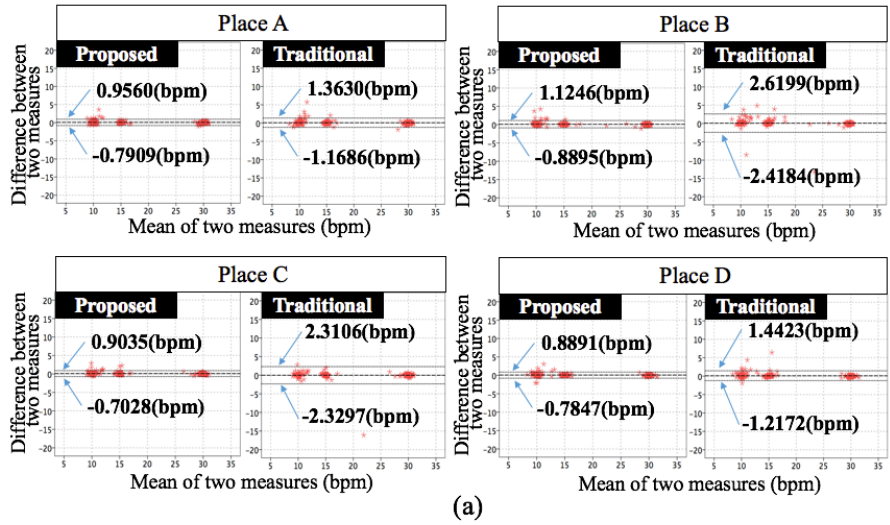

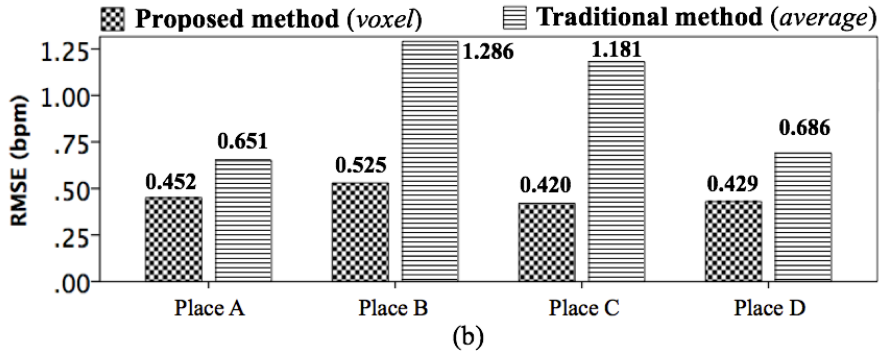

Fig. 12. Dataset 1 (separate results along with the different environment): (a) Bland-Altman plots of Thermal Voxel integration method and the traditional averaging method, and (b) RMSE comparisons for Place A – D.

*Dataset 2 and 3: Bland-Altman and RMSE analysis*

Given the signal quality of the ground truth measurements for Datasets 2 and 3 were much poorer than those for Dataset 1 (see Figure 10 (c,f,i)), the belt sensor could not act as the ground truth. Therefore, we tested the agreement by extracting the estimated respiration rates of which rSQI is greater than or equal to the mean from the ground truth in Dataset 1 (i.e. $P_r \geq 0.9825$). For Dataset 2, Thermal Voxel integration method produced a mean bias of 0.0650 bpm with 95% limits of agreement being -1.9591 to 2.0890 bpm (Figure 13(a)), while the traditional averaging method showed the mean bias of -0.2735 bpm with 95% limits of agreement being -4.4010 to 3.8540 bpm (Figure 13(b)). Accordingly, samples derived from Thermal Voxel integration method and the traditional approach were correlated with the reference with r=0.9579 (p<0.001) and r=0.8743 (p<0.001), respectively. The RMSE was also reduced more than twice from 2.11bpm (i.e. averaging method) to 1.03bpm (i.e. Thermal Voxel integration) (see Figure 13(c)). For Dataset 3, both methods produced less reliable results (Figure 14) (Thermal Voxel-based: mean bias = -0.1405 bpm, 95% limits of agreement: -4.9480 to 4.6670 bpm, r=0.8270 (p<0.001), RMSE=2.45bpm), the traditional one: mean bias=0.1921 bpm, 95% limits of agreement: -5.2682 to 4.8840 bpm, r=0.7990 (p<0.001), RMSE=2.59bpm).

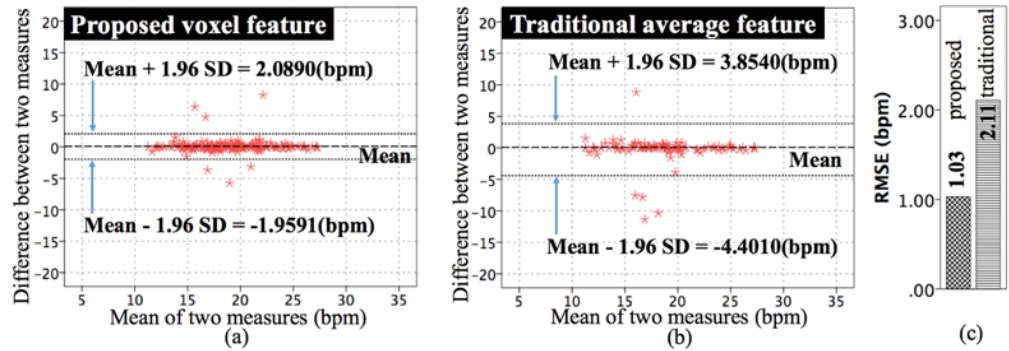

Fig. 13. Dataset 2: Bland-Altman plots of (a) Thermal Voxel integration method, (b) the traditional averaging based estimation method, and (c) overall RMSE comparisons. The mean $P_r$ value (rSQI) from the ground truth in Dataset 1 was set as exclusion criterion ($P_r \geq 0.9825$).

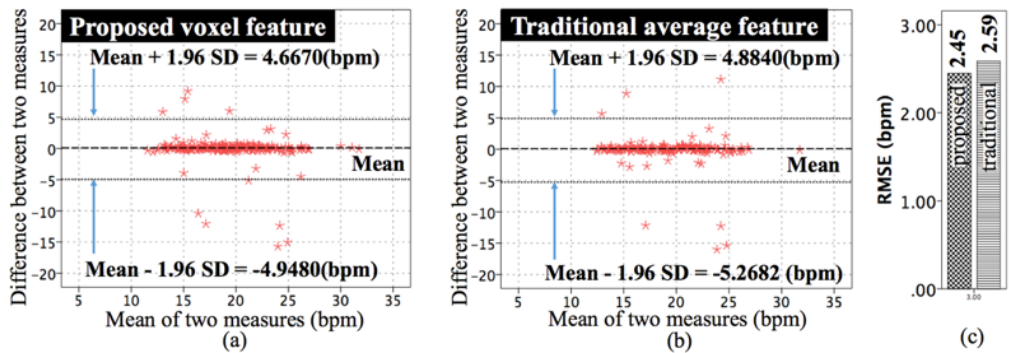

Fig. 14. Dataset 3: Bland-Altman plots of (a) Thermal Voxel integration method, (b) the traditional averaging method, and (c) overall RMSE comparisons. The mean $P_r$ value (rSQI) from the ground truth in Dataset 1 was set as exclusion criterion ($P_r \geq 0.9825$).

Finally, we illustrate the performance of nostril-ROI and respiration tracking algorithms together through examples in Visualization 1 (S1 in place C from Dataset 1), Visualization 2 (S9 from Dataset 2), and Visualization 3 (S3 from Dataset 3). Those visualizations (available online as supplementary material[1]) show the results for the state of the art algorithms (Nostril-ROI tracking: Median Flow, and respiration tracking: averaging method) and for our proposed three-stages approach.

## 6. Discussion

The robustness of thermal imaging in respiration measurements has been proven in previous research in indoor and stationary settings [6,10–13]. This paper extends the earlier research to explore how to accurately track respiration in ubiquitous situations with mobile thermography. To assess the robustness of our approach, we performed in-depth analysis of the experimental results through two evaluation steps – i) tracking of the nostril ROI and ii) tracking of the respiration rate.

Robust nostril-region tracking is critical to monitoring respiratory signs as discussed in [6,11]. In the latest work, to the best of our knowledge, Pereira *et al*. [6] achieved the highest performance in tracking the nostril region by adopting Mei *et al*.'s Sparse

---

[1] http://youngjuncho.com/2017/boe_supplementary/

Representation method [16] – one of the most advanced motion tracking algorithms – in controlled laboratory experiments. Outside the laboratory, however, key challenges had not been addressed - i.e. high thermal dynamic ranges and noises amplified by motions and breathing dynamics - which reduce tracking performance. To overcome these challenges, we proposed the Optimal Quantization algorithm to improve the dynamic range of the quantized signal, and the Thermal Gradient Flow algorithm to more robustly track the nostril ROI. This approach has been explored thoroughly by comparing its performances with both traditional (Mode seeking [30]) and state-of-the-art visual tracking algorithms (Sparse Representation [16] and Median Flow [21]).

In the case of environments with changes in ambient temperature with the participant maintaining a still posture (i.e. Dataset 1), our approach performed better than all the others. In particular, we found that the environment with high thermal dynamic ranges greatly influences the performance of the other methods. However, the Optimal Quantization approach meant that the Thermal Gradient Flow algorithm was unaffected. In scenes with low thermal dynamic ranges, on the other hand, both the recent advanced methods (i.e. Sparse Representation [16], Median Flow [21]) and Thermal Gradient Flow produced similarly almost perfect results, outperforming the earlier method (i.e. Mode Seeking [30]). Similarly, in the case of sedentary-motion scenario (i.e. Dataset 2), the three methods performed significantly better than the earlier approach without statistically significant differences between each other, despite the fact that our approach produces the highest performance. In the case of tracking during physical activity (i.e. Dataset 3: high thermal dynamic range & motion artifacts), however, our approach performed significantly better than all the other methods, showing its robustness in nostril-region tracking to challenges present in everyday settings (see Figure 8).

The conducted experiments do not cover all possible scenarios. These include extreme cases where there are sudden transitions between scenes with different levels of ambient temperature (e.g., leaving a heated building in winter) or under varying levels of humidity (e.g. swimming pool, sauna) which influence the temperature distribution [32]. Nonetheless, we expect that our tracking algorithm designed for thermal imaging sequences could help track other areas beyond the nostrils, which are less influenced by the breathing dynamics such as the nose tip and perioral regions where there are key thermal signatures in association with a person's psychological state [33].

The second evaluation study aimed to compare Thermal Voxel based approach with the temperature averaging method commonly used [6,10–13] for extracting one-dimensional respiration patterns. The comparison was run on the same nostril-ROI sequences which were fully tracked by the Optimal Quantization based Thermal Gradient Flow. The results show that the Thermal Voxel integration method generally outperformed traditional averaging approaches. In the body of work on the thermal imaging based respiration tracking, the step for the respiratory-feature extraction has by and large been overlooked by adopting the simple averaging method. We found that this new Thermal Voxel based approach is capable of improving the accuracy, even in controlled situations designed to simulate guided respiratory rates (i.e. Dataset 1, see Figure 11 and 12). It is noteworthy that the environments with wider range of ambient temperature tend to lessen the accuracy of the traditional temperature averaging method (e.g. Place B and C in Figure 12) while the Thermal Voxel integration method was relatively immune to these changes. Under motion artifacts (Dataset 2 and 3), both methods showed relatively low agreement with the chest-belt respiration sensor. It is possible that the thermal imaging-based measurement is not fully immune to a person's movement, even though our motion tracking method performs properly under motion artifacts. Nevertheless, the same applies to the chest-belt-based measurement (see Figure 10(h)), so we examined the signal quality of each measurement to exclude the less-reliable moments for comparison as in [5]. As discussed in [4], the goodness-based metric can be a replacement for signal-noise ratio if the observed signals has the periodic property of dominant frequency (i.e. respiration rate). Consequently, the rSQI was used to better compare both methods with the ground truth for Dataset 2 and 3.

Lastly, to support mobile situations, we employed the mobile thermal camera offering both the small-sized form and low computational-resource requirement for the recording. However, the spatial and temporal resolutions of this hardware system are relatively low in comparison with high-end thermographic systems. In addition, the temperature distribution sequences collected from the system include sporadically a few extreme values (e.g. over 100°C or lower than -30°C) wrongly calculated due to lens-inducing errors (e.g. lens focusing, misted edges of lens), which all could possibly degrade the performance in respiration tracking. Fortunately, the statistical-outlier reduction used in our Optimal Quantization technique is capable of minimizing negative effects of the latter issue. Other than this, the resolution-related aspects are less likely to be fundamentally improved by algorithmic approaches. We expect that the higher the spatial resolution of the data, the more enhanced the signals will be. Similarly, higher sampling rate can directly improve the accuracy of respiration indices such as the breathing variability as discussed in [4]. In addition, the performance of our approach may be affected by extremely fast breathing rate beyond the range of breathing rates observed in our experiments. This is due to the fact that the temporal resolution of the thermographic system we use depends on software schedulers of a mobile operating system (i.e. Android in our case) which leads to producing unsteady sampling rates. Therefore, it is important to note that thermal cameras with higher resolution can contribute to the further enhancement of respiration tracking performances.

## 7. Conclusion

In this paper, we have proposed a robust respiration tracking method using mobile thermal imaging. As a step forward to the deployment of thermal imaging as a practical respiration sensor, we have identified major challenges, such as high thermal dynamic range scenes and artifacts combined with breathing dynamics and motion, to be encountered in mobile situations. In addition, we have identified a weakness in the traditional respiration feature, i.e. average temperature over the nostril ROI, which produces weak signals due to the variance of viewing direction, shallow breathing and the low spatial resolution of mobile thermal imaging. To overcome the identified issues, we developed a novel *Optimal Quantization* method and a thermal gradient map-based visual tracking technique called *Thermal Gradient Flow* method. Furthermore, to strengthen the respiratory signal quality, we have introduced a novel *Thermal Voxel*-based integration method. With the designed experiments and their corresponding datasets, the performances of our methods in the tracking of the nostril region and the respiratory rate were evaluated. In particular, we have identified that this new nostril-tracking strategy performs significantly better in scenes with high thermal dynamic ranges compared to the three state-of-the-art algorithms. Highly robust accuracy in the tracking of respiratory rate was also achieved, suggesting the possibility to bring the thermal imaging based respiration monitoring to real-world situations. We expect that this new approach can be of help in a variety of other situations not investigated here, for example, to tailor the activity to a person's psychological needs (e.g. mental stress, anxiety). Initial results of the application of the methods to such application can be seen in [34].

## Funding

University College London Overseas Research Scholarship (UCL-ORS).